\documentclass{llncs}
\usepackage{amssymb}
\usepackage{times,amsmath,epsfig}
\usepackage{graphicx,amssymb}
\usepackage{helvet}
\usepackage{courier}
\usepackage{latexsym}
\usepackage[usenames]{color}
\usepackage{graphics}
\usepackage{enumerate,soul,xspace}
\definecolor{Orange}{rgb}{1.00,0.27,0.00}

\usepackage{url} %,theorem}

\newcommand{\sldots}{\mathit\mbox{{\tiny\ldots}}}

\newcommand{\vett}[1]{\vec{#1}}

\newcommand{\atom}[1]{\underline{#1}}

\newcommand{\dep}{\Sigma}

\newcommand{\isa}[1]{\mathit{ISA}}

\newcommand{\head}[1]{\mathit{head}(#1)}
\newcommand{\body}[1]{\mathit{body}(#1)}

\newcommand{\ans}[3]{\mathit{ans}(#1,#2,#3)}

\newcommand{\chase}[2]{\mathit{chase}(#1,#2)}

\renewcommand{\paragraph}[1]{\textbf{#1}}

\newcommand{\nc}{\newcommand}

\newcommand{\calC}{\ensuremath{{\cal C}}\xspace}
\newcommand{\bigO}{\ensuremath{{\cal O}}\xspace}

\newcommand{\dom}{\ensuremath{\text{dom}}\xspace}
\newcommand{\Neq}{\ensuremath{\text{Neq}}\xspace}

\newcommand{\Num}{\ensuremath{\text{Num}}\xspace}
\newcommand{\DNum}{\ensuremath{\text{DNum}}\xspace}
\newcommand{\IfThen}{\ensuremath{\text{IfThen}}\xspace}
\newcommand{\IfTheni}{\ensuremath{\text{IfThen2}}\xspace}
\newcommand{\IfThenii}{\ensuremath{\text{IfThen3}}\xspace}
\newcommand{\IfEq}{\ensuremath{\text{IfEq}}\xspace}
\newcommand{\NotB}{\ensuremath{\text{NotB}}\xspace}
\newcommand{\OrB}{\ensuremath{\text{OrB}}\xspace}
\newcommand{\TrueB}{\ensuremath{\text{TrueB}}\xspace}

\newcommand{\rgoal}{\ensuremath{r_{\text{goal}}}\xspace} % Name of the
\newcommand{\rtuples}{\ensuremath{r_{\text{tuples}}}\xspace} % Name of the
\newcommand{\rchaseok}{\ensuremath{r_{\text{chase}}}\xspace} % Name of the
\newcommand{\rquery}{\ensuremath{r_{\text{query}}}\xspace} % Name of the

\newcommand{\Ptuples}{\ensuremath{P_{\text{tuples}}}\xspace} % Name of the
\newcommand{\Pchase}{\ensuremath{P_{\text{chase}}}\xspace} % Name of the
\newcommand{\Pquery}{\ensuremath{P_{\text{query}}}\xspace} % Name of the

\newcommand{\serule}[3]{\ensuremath{ #2 \ra \exists #1\; #3}}  % tgd
\newcommand{\srule}[2]{\ensuremath{ #1 \ra  #2}}  % tgd
\newcommand{\srulelb}[2]{\ensuremath{ #1 \ra\\\mbox{ }\hfill  #2\\}}  % tgd with linebreak
\newcommand{\drule}[2]{\ensuremath{ #2 :-  #1.}}  % Datalog rule
\newcommand{\longdrule}[2]{\begin{multline*} #2 :-\\
      #1.\end{multline*}}  % multi line Datalog rule
\nc{\thomas}[1]{}
\nc{\Georg}[1]{}

\setul{}{0.3mm}
\setstcolor{red}
\newcommand{\oldnew}[3]{{\textcolor{blue}{\setlength{\fboxsep}{1pt}\fbox{\small #1}}} \st{#2} \textcolor{blue}{#3}}
\renewcommand{\oldnew}[3]{#3}

\nc{\gon}[2][]{\oldnew{G}{#1}{#2}} 
\nc{\ton}[2][]{\oldnew{T}{#1}{#2}}

\nc{\todo}[1]{}

\nc{\dlorfull}[2]{#2}
\newcommand{\A}{\mathcal{A}}

 \newcommand{\R}{\mathcal{R}}

\newcommand{\ra}{\rightarrow}

\newcommand{\dd}[2]{#1_1,\ldots,#1_{#2}}             % x1,...,xn (da da)
\def\qed{\hfill{\qedboxempty}      % qed with empty box
  \ifdim\lastskip<\medskipamount \removelastskip\penalty55\medskip\fi}

\def\qedboxempty{\vbox{\hrule\hbox{\vrule\kern3pt
                 \vbox{\kern3pt\kern3pt}\kern3pt\vrule}\hrule}}

\def\qedfull{\hfill{\qedboxfull}   % qed with full box
  \ifdim\lastskip<\medskipamount \removelastskip\penalty55\medskip\fi}

\def\qedboxfull{\vrule height 4pt width 4pt depth 0pt}

\newcommand{\markfull}{\qedboxfull}

\newcommand{\nop}[1]{}

\newcommand{\sol}[2]{\mathit{mods}(#1,#2)} 
\newcommand{\subseteqs}{\,{\subseteq}\,}
\newcommand{\ges}{\,{\ge}\,}

\newcommand{\cups}{\,{\cup}\,}
\nop{
\newtheorem{theorem}{Theorem}[chapter]
\newtheorem{corollary}[theorem]{Corollary}

\newtheorem{lemma}[theorem]{Lemma}

\newtheorem{definition}{Definition}
\newtheorem{example}{Example}
} % END NOP
\begin{document}
\urldef\arxiv\url{http://arxiv.org/abs/1106.3767}
\sloppy
\title{Rewriting Ontological Queries into Small Nonrecursive  Datalog Programs\thanks{\dlorfull{Future improvements and  
extended versions of this paper will be published in}{This paper extends a shorter version presented at  {\em DL 2011, 
24th International Workshop on Description Logics} by mainly expanding Section~\ref{sec:main}. Further 
extended and/or improved versions will be posted as they arise to} arXive-CORR
at 
%\mbox{\url{http://arxiv.org/abs/1106.3767}
\arxiv}}
\titlerunning{Rewriting Ontological Queries}
\toctitle{Rewriting Ontological Queries into Small Datalog Programs}
%\subtitle{Extended Abstract}
\author{Georg Gottlob\inst{1} and Thomas Schwentick\inst{2}}
\authorrunning{G.~Gottlob and T.~Schwentick}
\tocauthor{Georg Gottlob  and Thomas Schwentick}
\institute{Department of Computer Science, University of Oxford \email{gottlob@cs.ox.ac.uk} \\
\and Fakult\"at f\"ur Informatik, TU Dortmund \email{thomas.schwentick@udo.edu}}

\maketitle

\begin{abstract}
We consider the setting of ontological database access, where an A-box is given in form of a relational database $D$ and 
where a Boolean conjunctive query $q$  has to be evaluated against $D$ modulo a $T$-box $\Sigma$ formulated in 
DL-Lite or Linear Datalog$^\pm$.
 It is well-known that $(\Sigma,q)$ can be 
rewritten into an equivalent  nonrecursive Datalog program $P$ that can be directly evaluated over $D$. However, 
for Linear Datalog$^\pm$ or for DL-Lite versions that allow for  role inclusion, 
the rewriting methods 
described so far result in a nonrecursive Datalog program $P$ of size exponential in the joint size of $\Sigma$ and  $q$. This gives rise to the 
interesting question of whether such a rewriting necessarily needs to be of exponential size.
In this paper we show that 
it is actually possible to translate $(\Sigma,q)$ into a polynomially sized  equivalent nonrecursive Datalog program $P$. 
\end{abstract}

\section{Introduction}
\dlorfull{}{
\subsection{Motivation}
}
This paper is about query rewriting in the context of ontological database access. Query rewriting is an important new optimization 
technique specific to ontological queries. The essence of query 
rewriting, as will be explained in more detail below, is to compile a query and an ontological theory (usually formulated in some description logic or rule-based language) into a target query 
language that can be directly executed over a relational database 
management system (DBMS).
The advantage of such an approach is obvious. Query rewriting can be used as a preprocessing step 
for enabling the exploitation of mature and efficient existing 
database technology to answer ontological queries. In particular, 
after translating an ontological query into SQL, sophisticated query-optimization 
strategies can be used to efficiently answer it. However, there is a pitfall here. If the translation inflates the query excessively and creates from a reasonably sized ontological query an 
enormous exponentially sized SQL query (or SQL DDL program), 
then the best DBMS 
may be of little use. 

\nop{The above problem has motivated a florishing research activity at the cutting edge between the fields of description logic (DL) and 
database theory. While exprimental and commercial ontological database management systems have existed and have 
been used for a number of years (e.g.~\cite{Chong05,ACDL*05}), 
they have been problematic with respect to performance, 
and there is an agreement that ontological query optimization methods have not yet been developed to their full potential. 
In fact, there is a general feeling that many issues of 
ontology querying are not well understood, and 
that deep theoretical research is necessary to better understand some fundamental issues of query rewriting. In this spirit, 
the present paper tries to shed light on a relevant problem
in this context: Is it at all possible to translate an ontological 
conjunctive query (based on into a polynomially sized SQL (or SQL-DDL) query.} 

\dlorfull{
\paragraph{Main results.}
}
{
\subsection{Main Results}
}
%This paper is a theoretical one that attacks the above question.
We show that polynomially sized  query rewritings into nonrecursive
Datalog exist in specific settings. 
%GGNEW
Note that nonrecursive Datalog can be efficiently translated 
into SQL with view definitions (SQL DDL), which, in turn, can be directly 
executed over any standard DBMS.
 Our results 
are --- for the time being --- of theoretical nature and we do not 
claim that they will lead to better practical algorithms. 
This will be studied via implementations in the next future. 
Our main result applies to  the setting
 where ontological constraints are formulated in terms of {\em
   tuple-generating dependencies (tgds)}, and we make heavy use of the 
well-known {\em chase} procedure~\cite{MaMS79,JoKl84}. For
definitions, see Section~2. The result after chasing a tgd set
$\Sigma$ over 
a database $D$ is denoted by $chase(D,\Sigma)$. 

Consider a set $\Sigma$ of tgds  and a database $D$ over a joint signature $\cal R$. Let $q$ be a Boolean conjunctive  query (BCQ) 
issued against $(D,\Sigma)$. We would like to transform $q$
into a nonrecursive Datalog query $P$ such that 
$(D,\Sigma) \models q$ iff $D\models P$. 
 We assume here that $P$ has a special 
propositional goal $goal$, and 
$D\models P$ means that $goal$ is derivable 
 from $P$ when evaluated over $D$. 
Let us define an important property of classes of tgds.
\begin{definition}~\label{def:PWP}
{\bf Polynomial witness property (PWP).} The PWP 
holds for a class ${\cal C}$ of tgds if there exists a polynomial $\gamma$ such
that, for every finite set $\Sigma\subseteq {\cal C}$ of tgds 
and each 
BCQ $q$, the following holds: for each database $D$, whenever
$(D,\Sigma) \models q$, then there is a sequence of at most
\ton[$\gamma(|\Sigma|,|q|)$]{$\gamma(|\Sigma|,|q|)$}  chase steps
whose atoms already entail $q$. \thomas{Es erscheint mir natürlicher,
  wenn $\gamma$ zwei Parameter hat}
\end{definition}

Our main technical result, which is more formally stated 
\dlorfull{}
{
and proven
}
 in Section~\ref{sec:main},
is as follows. 
\dlorfull{}
{
\smallskip
}

\noindent
{\bf Theorem~1.} {\em Let $\Sigma$ be a set of tgds  from a class
  ${\cal C}$ enjoying the PWP.
Then each BCQ $q$ can be rewritten in polynomial 
time into a 
nonrecursive Datalog program $P$ of size polynomial in 
the joint size of $q$ and $\Sigma$, such that 
for every database $D$, $(D,\Sigma)\models q$ if
and only if $D\models P$. Moreover, the arity of $P$
is  $\max(a+1,3)$, where $a$ is 
the maximum arity of any predicate symbol occurring in $\Sigma$, 
in case a sufficiently 
large linear order can be accessed in the database, or
otherwise  
by $O(\max(a+1,3)\cdot\log m)$, where $m$ is the joint size of $q$ and $\Sigma$.}    

\dlorfull{
\paragraph{Other Results.}
}
{
\subsection{Other Results}
}
From this result, and from already established facts, a good number of further rewritabliity results for other formalisms can 
be derived. In particular, we can show that conjunctive queries 
based on other classes of tgds or description logics can 
be efficiently translated into nonrecursive Datalog. Among 
these formalisms are: linear tgds, originally defined in~\cite{CaGL09} and equivalent to inclusion dependencies, various major versions of the well-known
description logic DL-Lite~\cite{CDLL*07,PLCD*08}, and 
sticky tgds~\cite{CaGP11} as well as  sticky-join tgds~\cite{CaGP10a,CaGP10b}. 
\dlorfull{For space reasons, we }{We} will just give an overview and very 
short explanations of how each of these rewritability 
results follows from our main theorem. \dlorfull{}{A more detailed treatment is planned for a future
version of this paper.}  

\dlorfull{
\paragraph{Structure of the Paper.}
}
{
\subsection{Structure of the Paper}
}
The rest of the paper is structured as follows. In
Section~\ref{sec:prelim} we state a few preliminaries and simplifying
assumptions. In Section~\ref{sec:main}, we 
\dlorfull{
explain the idea of the proof of the main result.
}
{
give a rather detailed proof sketch of the main result.
}
Section~\ref{sec:further}, contains the 
other results following from the main result. A brief overview of
related work 
concludes the paper in Section~\ref{sec:related}.

\section{Preliminaries and Assumptions}\label{sec:prelim}
We assume the reader to be familiar with the terminology of relational 
databases and the concepts of {\em conjunctive query (CQ)} and {\em Boolean 
conjunctive query (BCQ)}. 
For simplicity, we restrict our attention to Boolean
conjunctive queries $q$. However, our results 
can easily be reformulated for queries with output, 
\dlorfull{see the 
extended version of this paper~\cite{extended})}{see Remark~3
after the proof of Theorem~\ref{theo:polyDatalog}}. 

%-------------------------------------------------------------
Given a relational schema~$\R$, a \emph{tuple-gene\-r\-ating dependency (tgd)} 
  $\sigma$ is a first-order formula of the form $\forall \vett{X}\forall \vett{Y}\,\Phi(\vett{X},$ $\vett{Y}) \,{\ra}\,
  \exists{\vett{Z}} \,\Psi(\vett{X},\vett{Z})$, where $\Phi(\vett{X},\vett{Y})$
  and $\Psi(\vett{X},$ $\vett{Z})$ are conjunctions of atoms over~$\R$, called
  the \emph{body} and the \emph{head} of $\sigma$, denoted $\body{\sigma}$ and
  $\head{\sigma}$, respectively. We usually omit the universal quantifiers in tgds.
  Such~$\sigma$ is satisfied in a 
  database~$D$ for $\R$ iff, whenever there exists a homomorphism $h$ that maps the atoms of
  $\Phi(\vett{X},\vett{Y})$ to atoms of $D$, there exists an extension $h'$ of
  $h$ 
  that maps the atoms 
  of~$\Psi(\vett{X},\vett{Z})$ to atoms of $D$. 
  All sets of tgds are finite here. \ton{We assume in the rest of the paper
  that every tgd has exactly one atom and at most one existentially
  quantified variable in its head. 
 %GGNEW
A set of tgds is in {\em normal form} if the head of 
each tgd consists of a single atom.
  It was shown in
 \cite[Lemma 10]{CaGK08} that
  every set $\Sigma$ of TGDs can be transformed into a set $\Sigma'$ in normal form 
  of size at most quadratic in $|\Sigma|$, such that 
  $\Sigma$ and $\Sigma'$ are equivalent 
  with respect to query answering. 
  The normal form transformation 
  shown in \cite{CaGK08} can be achieved
  in logarithmic space. It is, moreover, easy to see that this very simple transformation preserves the polynomial witness property.}

\todo{Check where the normal form was actually shown - or show how to
  do it}

  For a database
  $D$ for $\R$, and a set of tgds $\dep$ on $\R$, the set of \emph{models} of $D$ and $\dep$, denoted $\sol{D}{\dep}$,
  is the set of all (possibly infinite) databases $B$ such that (i) $D\subseteqs B$
  %$D$ can be homomorphically mapped to $B$ 
  and (ii)
  every $\sigma\,{\in}\,\dep$ is satisfied in $B$.
The set of \emph{answers} for a CQ $q$ to $D$
  and $\dep$, denoted $\ans{q}{D}{\dep}$, is the set of all tuples
  $\atom{a}$ such that $\atom{a} \in
  q(B)$ for all $B \,{\in}\, \sol{D}{\dep}$.
  The {\em answer}~for a BCQ~$q$ to $D$ and $\dep$ is {\em yes}  iff
  the empty tuple is in $\ans{q}{D}{\dep}$, also denoted as
  $D\cup \Sigma\,{\models}\, q$.

Note that, in general, 
query answering under tgds is
undecidable~\cite{BeVa81}, 
even when the schema and tgds are fixed~\cite{CaGK08}. 
Query answering is, however, decidable for interesting classes of tgds, 
among which are those considered in the present paper.

The \emph{chase} procedure \dlorfull{~\cite{MaMS79,JoKl84}}{
was introduced 
to enable checking
implication of dependencies~\cite{MaMS79}, and later also for checking query
containment~\cite{JoKl84}.
It is a procedure for repairing a database relative to 
a set of 
dependencies, so that the result of the chase
satisfies the dependencies.  By 
``chase'', we refer
both to the chase procedure and to its output.}
\dlorfull{uses the following \emph{oblivious} chase rule. }{
The chase comes
in two flavors: \emph{restricted} and \emph{oblivious}, where the restricted chase
one applies tgds only when they are not satisfied (to repair them), while the oblivious chase 
always applies tgds (if they produce a new result).  We focus on 
the oblivious one, since it makes proofs technically simpler.
The \emph{(oblivious) tgd chase rule} defined below is the building block 
of the chase.
}

\textsc{TGD Chase Rule.}  Consider a data\-base $D$ for a
relational schema $\R$, and a tgd~$\sigma$ on $\R$ of the form $\Phi(\vett{X},\vett{Y})
\ra \exists{\vett{Z}}\,  \Psi(\vett{X},$ $\vett{Z})$.  Then, $\sigma$ is \emph{applicable} to~$D$ if
there exists a homomorphism $h$ that maps the atoms of $\Phi(\vett{X},\vett{Y})$ to
atoms of~$D$.  Let $\sigma$ be applicable to $D$, and $h_1$ be a homomorphism that
extends $h$ as follows: for each $X_i\in \vett{X}$, $h_1(X_i)=h(X_i)$; for each
$Z_j\in \vett{Z}$, $h_1(Z_j)=z_j$, where $z_j$ is a fresh  null value
(i.e., a Skolem constant)
different from all 
nulls already introduced. 
The {\em application of} $\sigma$ on $D$ adds 
to~$D$ 
the atom $h_1(\Psi(\vett{X},\vett{Z}))$ if 
not already
in~$D$ (which is possible when $\vett{Z}$ is empty).~\markfull

The chase algorithm for a database $D$ and a set of tgds $\Sigma$ consists of an exhaustive application of the tgd chase
rule in a breadth-first (level-sat\-urating) fashion,  
which leads as result to a (possibly infinite) chase for $D$
and~$\Sigma$.
\dlorfull{Each atom from the database $D$
is assigned a {\em derivation level}. Atoms in $D$ have derivation level $0$. 
If an atom has not already derivation level $\leq i$ but can be obtained by a single  application of a tgd via the chase rule 
from atoms having derivation level $\leq i$, then its derivation 
level is $i+1$. The set of all 
atoms of derivation level $\leq k$ is denoted by
$\mathit{chase}^{k}(D,\Sigma)$.}{
Formally, 
the {\em chase of level up to~$0$} of $D$ relative to~$\Sigma$, de\-no\-ted~$\mathit{chase}^0(D,\Sigma)$, is defined as $D$, 
assigning to every atom in $D$ the \emph{(derivation) level} $0$. 
For every $k\ges 1$, the~{\em chase of level up to~$k$} of~$D$ 
relative to~$\Sigma$, de\-no\-ted~$\mathit{chase}^k(D,\Sigma)$, is constructed as follows: 
let $\dd{I}{n}$ be all possible images of bodies of tgds in $\dep$ relative to some
homomorphism such that (i) $\dd{I}{n}\subseteqs \mathit{chase}^{k-1}(D,\Sigma)$ 
and (ii)~the highest level of an atom in some
$I_i$ is $k-1$; %among   
%, and $\atom{a}_i$ be the atom with highest level in $I_i$;
%let $M$ be such that $\level{\atom{a}_M} = \min_{1 \leq i \leq k}
%\{\level{\atom{a}_i}\}$: 
then, perform every corresponding tgd application on $\mathit{chase}^{k-1}(D,\Sigma)$, choosing the applied tgds and homomorphisms in a linear and 
lexicographic order, respectively, and assigning to every new atom the \emph{(derivation) level} $k$.}
%    For brevity, the application of the chase rule with a tgd $\sigma$ 
%    on $D$ is called an application of~$\sigma$~on~$D$.
%The {\em chase of level up 
%to~$k\ge 0$} of $D$ relative to~$\Sigma$, de\-no\-ted~$\mathit{chase}^k(D,\Sigma)$, 
%is the set of all atoms in $\chase{D}{\dep}$ of level at most~$k$.
The {\em chase} of $D$ relative to $\Sigma$, denoted $\chase{D}{\dep}$, is\dlorfull{~then}{~thus}  the limit 
of $\mathit{chase}^k(D,\Sigma)$ for $k\rightarrow\infty$.

The (possibly infinite) chase relative to tgds
is a \emph{universal mo\-del}, 
i.e., there exists a homomorphism from $\chase{D}{\dep}$ onto every 
$B \,{\in}\,
\sol{D}{\dep}$~\cite{DeNR08,CaGK08}. This result implies that BCQs $q$ over $D$ and $\Sigma$ can be evaluated on the chase 
for~$D$ and $\Sigma$, i.e., $D\cups \Sigma \models q$ is equivalent to $\chase{D}{\dep}\models q$.

A {\em chase sequence} of length $n$ based on $D$ and $\Sigma$ 
is a sequence of $n$ atoms such that each atom is either from 
$D$ or can be derived via a single application of some 
rule in $\Sigma$ from previous atoms in the sequence.
If $S$ is such a chase sequence and $q$ a conjunctive query, 
we write $S\models q$ if there is a homomorphism from
$q$ to the set of atoms of $S$.

%-------------------------------------------------------------

We assume that every database has two constants,   $0$ and $1$,
that are available via the unary predicates 
$Zero$ and $One$, respectively. Moreover, each database has a 
binary predicate $\Neq$ such that $\Neq(a,b)$ is true precisely if 
$a$ and $b$ are distinct values.

We finally define {\em   $N$-numerical databases}. 
\ton
[For a
  database $D$ we write $D_N$ for the database obtained from $D$]
{Let
  $D$ be a database whose domain does not contain any natural
  numbers. We define $D_N$ as the extension of $D$}
by adding the natural numbers $0,1,\ldots,N$ to its domain, a unary relation
$\Num$ that contains exactly the numbers  $1,\ldots,N$, binary
order relations $Succ$ and 
$<$ on $0,1,\ldots,N$, expressing the natural successor and ``$<$'' orders 
on $N$, respectively. \footnote{Of course, if $\dom(D)$ already
  contains some natural numbers we can add a fresh copy of
  $\{0,1,\ldots,N\}$ instead.}
\ton[(Here we assume that the classical data values 
in $\dom(D)$ are 
all non-numeric.)]
{}
%
%and a unary relation $\Num_i$, for each $i\le N$
%interpreted as $\{i\}$. 
We refer to $D_N$ as 
the $N$-\emph{numerical
  extension} of $D$, and, a so extended database as \emph{$N$-numerical
  database}. 
\ton{We denote the total domain of a numerical database $D_N$
by $\dom_N(D)$ and the non-numerical 
domain (still) by $\dom(D)$. 
} Standard databases 
  can always be considered to be $N$-numerical, for some large $N$ by
  the standard type \emph{integer}, with the $<$ predicate (and even
  arithmetic operations). A number $maxint$ corresponding to 
 $N$ can be defined.

\section{Main Result}\label{sec:main}
  
Our main result is more formally stated as follows:

\begin{theorem} \label{theo:polyDatalog} 
 
Let ${\cal C}$ be a class of tgds in normal form, enjoying the  polynomial
  witness property and let $\gamma$ be the polynomial bounding the
  number of chase steps (with $\gamma(n_1,n_2)\ge \max(n_1,n_2)$, for all naturals $n_1,n_2$).
For each set  $\Sigma\subseteq{\cal C}$  of tgds   and  each Boolean
  CQ $q$, one can compute in polynomial time a nonrecursive Datalog program $P$ of
  polynomial size in $|\Sigma|$ and $|q|$,  such that, for every
  database $D$ it holds $D,\Sigma\models q$ if
    and only if $D\models P$. Furthermore:
    \begin{enumerate}
    \item[(a)] For $N$-numerical databases $D$, where $N \ge
      \gamma(|\Sigma|,|q|)$,   the arity of $P$  is  $\max(a+1,3)$,
      where $a$ is 
the maximum arity of any predicate symbol occurring in $\Sigma$;
    \item[(b)] otherwise (for non-numerical databases), 
    the arity of $P$ is $\bigO(\max(a+1,3)\cdot\log \gamma(|\Sigma|,|q|))$, 
%GGNEW Hallo, was ist k da unten?
%    where $k$ and $a$ are as above.
where $a$ is as above.
    \end{enumerate}
\end{theorem}

We note that $N$ is polynomially bounded in $|\Sigma|$ and $|q|$ by
the polynomial $\gamma$ that only depends on ${\cal C}$.
\dlorfull{
The rest of this section explains the basic ideas of the proof of this
result. A more detailed proof is given in~\cite{extended}.

\todo{Add ArXive number if available}
}
{
\smallskip

The rest of this section is dedicated to a detailed proof of 
Theorem~\ref{theo:polyDatalog}. 
}

\dlorfull{}{\smallskip}

\paragraph{\dlorfull{High-level idea of the proof}{Proof}.}
We first describe the \dlorfull{high level idea of the}{}  construction of a Datalog program
$P$ of arity $a+k+4$, where $k$ is the maximum number of tuples in any
left hand side of a chase rule. We explain afterwards, how the arity
can be reduced to $\max(a+1,3)$. The program $P$  checks whether there is a
chase sequence $S=t_1,\ldots,t_N$ with
respect to $D$ and $\Sigma$ and a homomorphism $h$ from
$q$ to (the set of atoms of) $S$. 
To this end, $P$  consists of one
large rule \rgoal of polynomial size in $N$ and some shorter
rules that define auxiliary relations and will be explained below. 

The aim of \rgoal is to guess the chase sequence $S$ \emph{and} the
homomorphism $q$ at the same time. We recall that $N$ does not depend
on the size of $D$ but only on $|\Sigma|$ and $|q|$ and thus \rgoal
can well be as long as the chase sequence and $q$ together. One of the
advantages of this approach is that we only have to deal with 
those null values that are actually relevant for answering the
query. Thus, at most $N$ null values need to be represented.

One might
try to obtain \rgoal by just taking one atom $A_i$ for each tuple $t_i$ of
$S$ and one atom for each atom of $q$ and somehow test that they are
consistent. However, it is not clear how consistency could
possibly be checked in a purely conjunctive
fashion.\footnote{Furthermore, of course, there are no relations to
  which the atoms $A_i$ could possible be matched.} There are two ways
in which disjunctive reasoning is needed.
First, it is not a priori clear on which previous tuples, tuple
  $t_i$ will depend.  Second, it is not a priori clear to which tuples
  of $S$ the atoms of $q$ can be mapped. 

To overcome these challenges we use the following basic ideas.
\begin{enumerate}[(1)]
\item We represent the tuples of $S$ (and the required tuples of $D$)
  in a symbolic fashion, utilizing the numerical domain. 
\item We let $P$ compute auxiliary predicates that allow us to express
  disjunctive relationships between the tuples in $S$.
\end{enumerate}

\begin{example}\label{ex:a}
We illustrate the proof idea with a very simple running example, shown in
Figure~\ref{fig:exa}. 
\end{example}
\begin{figure}[h]
\centering
\begin{minipage}{7.5cm}
 
  \begin{enumerate}[(a)]
  \item $\Sigma:$
    \begin{enumerate}[$\sigma_1$:]
    \item \serule{Z}{R_1(X,Y)}{R_4(X,Y,Z)}
    \item 
\serule{X}{R_2(Y,Z)}{R_4(X,Y,Z)}
    \item
 \serule{Y}{R_3(X,Z)}{R_4(X,Y,Z)}
    \item 
\srule{R_4(X_1,Y_1,Z_1),R_4(X_2,Y_2,Z_2) }{ R_5(X_1,Z_2)}
   \end{enumerate}
  \end{enumerate}
\end{minipage}
\begin{minipage}[h]{4cm}
  \begin{itemize}
  \item[(b)] $q:$ $R_5(X,Y), R_3(Y,X)$
  \item[(c)]  $D:$
%\centerline{
$  
\begin{array}[t]{|c|c|}
    \hline
\multicolumn{2}{|c|}{R_1} \\\hline
 a & b \\
 c & d\\\hline
  \end{array}
\hspace{5mm}
\begin{array}[t]{|c|c|}
    \hline
\multicolumn{2}{|c|}{R_2} \\\hline
 e & g\\\hline
  \end{array}
\hspace{5mm}
\begin{array}[t]{|c|c|}
    \hline
\multicolumn{2}{|c|}{R_3} \\\hline
 g & a \\
 g & h\\\hline
  \end{array}
$
%}
  \end{itemize}
\end{minipage}

  \caption{Simple example with (a) a set $\Sigma$ of tgds, (b) a query
    $q$ and (c) a database $D$.}
  \label{fig:exa}
\end{figure}
A possible chase sequence in this example is shown in Figure
\ref{fig:chaseenc}(a). The mapping $X\mapsto a$ and
$Y\mapsto g$,  maps $R_5(X,Y)$ to $t_5$ and $R_3(Y,X)$ to $t_6$, thus
satisfying $q$.
\begin{figure}[h]
\footnotesize
  \begin{minipage}[h]{3.5cm}
(a)
\begin{itemize}
\item $t_1$: $R_1(a,b)$
\item $t_2$: $R_4(a,b,\bot_2)$
\item $t_3$: $R_2(e,g)$
\item $t_4$: $R_4(\bot_4,e,g)$
\item $t_5$: $R_5(a,g)$
\item $t_6$: $R_3(g,a)$
\end{itemize}    
  \end{minipage}
  \begin{minipage}[h]{3.5cm}
(b) 
   \begin{itemize}
\item $t_1$: $R_1(a,b,a)$
\item $t_2$: $R_4(a,b,\bot_2)$
\item $t_3$: $R_2(e,g,e)$
\item $t_4$: $R_4(\bot_4,e,g)$
\item $t_5$: $R_5(a,g,a)$
\item $t_6$: $R_3(g,a,g)$
\end{itemize}
  \end{minipage}
\hfill
  \begin{minipage}[h]{4.5cm}
(c)
$
  \begin{array}[h]{|*{9}{c|}}
    \hline
i & r_i & f_i & x_{i1} & x_{i2} & x_{i3} & s_{i} & c_{i1} &
c_{i2}\\\hline
1 & 1 & 0 & a & b & a & 0 & 0 & 0 \\
2 & 4 & 1 & a & b & 2 & 1 & 1 & 1 \\
3 & 2 & 0 & e & g & e & 0 & 0 & 0 \\
4 & 4 & 1 & 4 & e & g & 2 & 3 & 3 \\
5 & 5 & 1 & a & g & a & 4 & 2 & 4 \\
6 & 3 & 0 & g & a & g & 0 & 0 & 0 \\\hline
  \end{array}
$
    
  \end{minipage}

  \caption{(a) Example chase sequence, (b) its extension and (c) its
    encoding. $t_2$ is obtained by applying $\sigma_1$ to
    $t_1$. Likewise $t_4$ and $t_5$ are obtained by applying
    $\sigma_2$ to $t_3$ and $\sigma_4$ to $t_2$ and $t_4$, respectively.}
  \label{fig:chaseenc}
\end{figure}
\dlorfull{}{
Before we describe the proof idea in more detail, we fix some notation
and convenient conventions.
}

\paragraph{Notation and conventions.}
%
%Let $\cal R$ be a fixed signature with (a fixed number $r$ of) predicate symbols 
Let $\calC$ be a class of tgds enjoying the PWP,
let $\Sigma$ be a set of tgds from $\calC$, and let $q$ be a BCQ.
Let $R_1,\ldots R_m$ be the predicate symbols occurring in 
$\Sigma$ or in $q$. We denote the number of tgds in $\Sigma$ by
$\ell$.
  
Let $N:=\gamma(|\Sigma|,|q|)$ where $\gamma$ is as 
in Definition~\ref{def:PWP}, thus  $N$ is polynomial in 
$|\Sigma|$ and $|q|$. By definition of $N$, if $(D,\Sigma)\models q$, then
$q$ can be witnessed by a chase sequence $\Gamma$ of length $\leq
N$. Our assumption that $\gamma(n_1,n_2)\ge \max(n_1,n_2)$, for every
$n_1,n_2$, guarantees
that 
$N$ is larger than (i) the number of predicate symbols occurring in $\Sigma$, 
(ii) the cardinality $|q|$ of the query, and (iii) 
the number of rules in $\Sigma$.

% TS: neue Formulierung und etwas nach hinten geschoben
 
For the sake of a simpler presentation, we 
assume that all
relations in $\Sigma$ have the same arity $a$ and all rules use the
same number $k$ of tuples in their body. The latter can be easily
achieved by repeating tuples, the former by filling up shorter tuples
by repeating the first tuple entry.  Furthermore, we only consider chase
sequences of length $N$. Shorter sequences can be extended by adding tuples
from $D$.
\begin{example}\label{ex:b}
 Example \ref{ex:a} thus translates as illustrated in Figure \ref{fig:exb}.
 The (extended) chase sequence is shown in Figure \ref{fig:chaseenc} (b).
The query $q$ is now satisfied by the mapping $X\mapsto a$,
$Y\mapsto g$, $U\mapsto g$, $V\mapsto a$, thus mapping $R_5(X,Y,X)$ to
$t_5$ and $R_3(Y,X,Y)$ to $t_6$.
\end{example}
\begin{figure}[h]
\centering
\begin{minipage}[h]{7.5cm}
 \footnotesize
  \begin{enumerate}[(a)]
  \item $\Sigma:$
     \begin{enumerate}[$\sigma_1$:]
    \item \serule{Z}{R_1(X,Y,X),R_1(X,Y,X)}{R_4(X,Y,Z)}
    \item 
\serule{X}{R_2(Y,Z,Y),R_2(Y,Z,Y)}{R_4(X,Y,Z)}
    \item
 \serule{Y}{R_3(X,Z,X),R_3(X,Z,X)}{R_4(X,Y,Z)}
    \item 
\srulelb{R_4(X_1,Y_1,Z_1),R_4(X_2,Y_2,Z_2)}{R_5(X_1,Z_2,X_1)}
   \end{enumerate}
   \end{enumerate}
\end{minipage}
\begin{minipage}[h]{4.5cm}
  \begin{itemize}
  \item[(b)] $q:$ $R_5(X,Y,U), R_3(Y,X,V)$
  \item[(c)]
$D:$
%\centerline{
$  
\begin{array}[t]{|c|c|c|}
    \hline
\multicolumn{3}{|c|}{R_1} \\\hline
 a & b & a\\
 c & d & c\\\hline
  \end{array}
\hspace{2mm}
\begin{array}[t]{|c|c|c|}
    \hline
\multicolumn{3}{|c|}{R_2} \\\hline
 e & g& e\\\hline
  \end{array}
\hspace{2mm}
\begin{array}[t]{|c|c|c|}
    \hline
\multicolumn{3}{|c|}{R_3} \\\hline
 g & a & g\\
 g & h & g\\\hline
  \end{array}
$
%}
 \end{itemize}
\end{minipage}

  \caption{Modified example with (a) a set $\Sigma$ of tgds, (b) a query
    $q$ and (c) a database $D$.}
  \label{fig:exb}
\end{figure}

\paragraph{Proof \dlorfull{idea}{} (continued).}
On an abstract level, the atoms that make up the final rule \rgoal of
$P$ can be divided into three groups serving three different
purposes. That is, \rgoal can be considered as a conjunction
$\rtuples\land \rchaseok \land\rquery$. Each group is ``supported'' by
a sub-program of $P$ that defines relations that are used in \rgoal,
and we refer to these three subprograms as 
\Ptuples, \Pchase and \Pquery, respectively.
\begin{itemize}
\item The purpose of \rtuples is basically to lay the ground
for the other two. It consists of $N$ atoms that allow to guess the
symbolic encoding of a sequence $S=t_1,\ldots,t_N$.
\item The atoms of \rchaseok are
designed to verify that $S$ is an actual chase sequence with respect
to $D$.
\item Finally, \rquery checks that there is a homomorphism from $q$ to $S$.
\end{itemize}

\paragraph{\Ptuples and \rtuples.}
\dlorfull{}
{
We continue with an explanation of the symbolic representation of
tuples underlying \rtuples.

}
The symbolic representation of the tuples $t_i$ of the chase sequence
$S$ uses numerical values 
to encode null values, predicate symbols $R_i$
(by $i$), tgds 
$\sigma_j\in\Sigma$ (by $j$) and the number of a tuple $t_i$ in the
sequence (that is: $i$). 

In particular, the symbolic encoding  uses
the following numerical parameters.\footnote{We  use the names of the
  parameters as variable names in \rgoal as well.}
\begin{itemize}
\item $r_i$ to indicate the relation $R_{r_i}$ to which the tuple belongs;
\item $f_i$ to indicate whether $t_i$ is from $D$ ($f_i=0$ ) or
  yielded by the chase (
  $f_i=1$);
\item Furthermore, $x_{i1},\ldots,x_{ia}$ represent the attribute
  values of $t_i$ as follows. If the $j$-th attribute of $t_i$ is a value from $\dom(D)$ then
  $x_{ij}$ is intended to be that value, otherwise it is a null
  represented by a
  numeric value. 
\end{itemize}
Since each rule of $\Sigma$ has at most one existential 
quantifier in its head, at each chase step, at most one new null value 
can be introduced. Thus, we can unambiguously represent the null value
(possibly) introduced in the $j$-th step of the chase by the number $j$.
\dlorfull{}
{
 In particular, all
null values introduced in a chase sequence (of length $N$) can indeed
be represented by elements of the numerical domain. 
}

The remaining parameters $s_i$ and $c_{i1},\ldots,c_{ik}$ are used to
encode information about the tgd and the tuples (atoms) in $S$ that are used to generate the current tuple. More precisely,
\dlorfull{$s_i$ is intended to be the
number of the applied tgd $\sigma_{s_i}$ and $c_{i1},\ldots,c_{ik}$
are the tuple numbers of the $k$ tuples that are used to yield
$t_i$.
}
{
\begin{itemize}
\item $s_i$ is intended to be the
number of the applied tgd $\sigma_{s_i}$ and
\item $c_{i1},\ldots,c_{ik}$
are the tuple numbers of the $k$ tuples that are used to yield
$t_i$.
\end{itemize}
}
In the example, e.g., $t_5$ is obtained by applying $\sigma_4$
to $t_2$ and $t_4$.
The encoding of our running example can be found in Figure
\ref{fig:chaseenc} (c). 

We use a new relational symbol $T$ of arity $a+k+4$  not present
in the schema of $D$ for the representation of the tuples from
$S$. Thus, \rtuples is just:\\
%Indizes falsch! 
%$T(1,r_1,f_1,x_{i1},\ldots,x_{ia},s_i,c_{i1},\ldots,c_{ik}),\ldots,
%T(N,r_N,f_N,x_{iN},\sldots,x_{ia},s_i,c_{iN},\sldots,c_{ik})$. 
$T(1,r_1,f_1,x_{11},\ldots,x_{1a},s_1,c_{11},\ldots,c_{1k}),\sldots,$\\ \phantom{.} \hfill $T(N,r_N,f_N,x_{N1},\ldots,x_{Na},s_N,c_{N1},\ldots,c_{Nk})$.

The sub-program  $\Ptuples$ is intended to ``fill'' $T$ with suitable
tuples. 
\dlorfull{Basically, $T$ contains all encodings of tuples in $D$ (with
  $f_i=0$) and all syntactically meaningful tuples corresponding to
  possible chase steps (with $f_i=1$). 
}
{
The intention is that $T$ contains all
tuples that could be used in a chase sequence in principle. At this
point, there are no restrictions regarding the chase rules.
To this end, $\Ptuples$ uses two kinds of
rules, one for tuples from $D$ and one for tuples yielded by
the chase.
For each relation symbol $R_j$ of $D$, $\Ptuples$ has a rule\\

%\begin{itemize}
%\item 
\noindent
\drule{R_j(X_1,\ldots,X_a),\Num(Z)}{T(Z,j,0,X1,\ldots,Xa,0,0,\ldots,0)}\\  
%\end{itemize}

which adds all tuples from $R_j$ to $T$ and makes them accessible for
every possible position ($Z$) in $S$.

The following rule adds tuples that can possibly be obtained by chase
steps.
%\begin{itemize}
%\item 
\begin{multline}
  \label{eq:1}
  T(Z,Y,1,X_1,\ldots,X_a,V,U_1,\ldots,U_k) :-\\
\Num(Z),\Num(Y),\DNum(X_1),\ldots,\DNum(X_a),\\\Num(V),\Num(U_1),\ldots,\Num(U_k),\\1\le Y\le
  m,1\le V\le \ell, U_1<Z,\ldots,U_k<Z
\end{multline}
%\end{itemize}

Here, the first two inequalities make sure that only allowed relation
and tgd numbers are used, the latter inequalities guarantee that
to yield a tuple by a chase rule only tuples with smaller numbers can
be used.\footnote{As the latter constraints are independent from the
  concrete tgds, we decided to put them here. They could as well be
  tested in \rchaseok.} The rule uses one further predicate \DNum that
has not yet been defined. Its purpose is to contain all possible
values, that is: $\dom(D)\cup\Num$. It is (easily) defined by further
rules of \Ptuples. Note that this leaves the values for the
$X_j$ unconstrained, hence they can carry either domain values or
numerical values.
}

\dlorfull{
\paragraph{\Pchase and \rchaseok.} The following kinds of conditions have to be checked to ensure that
the tuples ``guessed'' by \rtuples constitute a chase sequence.
\begin{enumerate}[(1)]
\item For every $i$, the relation $R_{r_i}$ of a tuple $t_i$ has to match the head of its rule
  $\sigma_{s_i}$.
  \begin{itemize}
  \item In the example, e.g., $r_4$ has to be $4$ as the head of
    $\sigma_2$ is an $R_4$-atom.
  \end{itemize}
\item Likewise, for each $i$ and $j$ the relation number of tuple
  $t_{c_{ij}}$ has to be the relation number of the $j$-th atom of $\sigma_{s_i}$.
  \begin{itemize}
  \item In the example, e.g., $r_2$ must be $4$, as $c_{5,1}=2$ and
    the first atom of $\sigma_{s_5}=\sigma_4$ is an $R_4$-atom.
  \end{itemize}
\item If the head of $\sigma_{s_i}$ contains an
  existentially quantified variable, the new null value is represented
  by the numerical value $i$.
  \begin{itemize}
  \item This is illustrated by $t_4$ in the example: the first
    position of the head of rule 2 has an existentially quantified
    variable and thus $x_{4,1}=4$.
  \end{itemize}
\item If a variable occurs at two different positions  in $\sigma_{s_i}$
  then the corresponding positions in the tuples used to produce $t_i$
  carry the same value.
\item If a variable in the body of $\sigma_{s_i}$ also occurs in the head of
  $\sigma_{s_i}$ then the values of the corresponding positions in the body tuple and in
  $t_i$ are equal.
  \begin{itemize}
  \item $Z_2$ occurs in position 3 of the second atom of the body of $\sigma_4$ and
    in position 2 of its head. Therefore, $x_{4,3}$ and $x_{5,2}$ have
    to coincide (where the 4 is determined by $c_{5,2}$.
  \end{itemize}
\end{enumerate}
It turns out that all these tests can be done by \rchaseok, given
some relations that are precomputed by \Pchase. More precisely, we let \Pchase specify a 4-ary
predicate $\IfThen(X_1,X_2,U_1,U_2)$ that is intended to contain all tuples
fulfilling the condition: if $X_1=X_2$ then $U_1=U_2$. Similar
predicates are defined for conditions with two and three conjuncts in
the IF-part. Their definition by Datalog rules is straightforward.
}
{
\paragraph{\Pchase and \rchaseok.} Next, we describe the part of
\rgoal that checks that $S$ constitutes and actual chase sequence and
the rules of $P$ that specify the corresponding auxiliary relations.

The following kinds of conditions have to be checked to ensure that
the tuples ``guessed'' by \rtuples constitute a chase sequence.
\begin{enumerate}[(1)]
\item For every $i$, the relation $R_{r_i}$ of a tuple $t_i$ has to match the head of its rule
  $\sigma_{s_i}$.
  \begin{itemize}
  \item In the example, e.g., $r_4$ has to be $4$ as the head of
    $\sigma_2$ is an $R_4$-atom.
  \end{itemize}
\item Likewise, for each $i$ and $j$ the relation number of tuple
  $t_{c_{ij}}$ has to be the relation number of the $j$-th atom of $\sigma_{s_i}$.
  \begin{itemize}
  \item In the example, e.g., $r_2$ must be $4$, as $c_{5,1}=2$ and
    the first atom of $\sigma_{s_5}=\sigma_4$ is an $R_4$-atom.
  \end{itemize}
\item If the head of $\sigma_{s_i}$ contains an
  existentially quantified variable, the new null value is represented
  by the numerical value $i$.
  \begin{itemize}
  \item This is illustrated by $t_4$ in the example: the first
    position of the head of rule 2 has an existentially quantified
    variable and thus $x_{4,1}=4$.
  \end{itemize}
\item If a variable occurs at two different positions  in $\sigma_{s_i}$
  then the corresponding positions in the tuples used to produce $t_i$
  carry the same value.
\item If a variable in the body of $\sigma_{s_i}$ also occurs in the head of
  $\sigma_{s_i}$ then the values of the corresponding positions in the body tuple and in
  $t_i$ are equal.
  \begin{itemize}
  \item $Z_2$ occurs in position 3 of the second atom of the body of $\sigma_4$ and
    in position 2 of its head. Therefore, $x_{4,3}$ and $x_{5,2}$ have
    to coincide (where the 4 is determined by $c_{5,2}$.
  \end{itemize}
\end{enumerate}

Note that all these conditions depend on the given tgds. Indeed, every
tgd from $\Sigma$ contributes conditions of each of the five
forms. For the sake of simplicity of presentation, we explain the
effect of a tgd through the following example tgd that contains all
relevant features that might arise in a tgd. The generalization to
arbitrary tgds is straightforward but tedious to spell out in full
detail. Let us thus assume that $\sigma_1$ is the tgd\footnote{This
  example tgd is not related to our running example as that does not
  have a single tgd with all features.}
\[
\serule{V}{R_2(X,Y),R_3(Y,Z)}{R_4(X,V)}.  
\]
Condition (1) states that if a tuple $t_i$ is obtained by applying
$\sigma_1$ it should be a tuple from $R_4$. In terms of variables this
means, that for every $i$ it should hold: if $s_i=1$ then
$r_i=4$. 

This is the first occasion where we need some way to express a
disjunction in \rgoal (namely: $s_i\not=1 \lor r_i=4$). We can
meet this challenge with the help of an additional predicate to be
specified in \Pchase. More precisely, we let \Pchase specify a 4-ary
predicate $\IfThen(X_1,X_2,U_1,U_2)$ that is intended to contain all tuples
fulfilling the condition: if $X_1=X_2$ then $U_1=U_2$. 
\IfThen can be specified by the following two rules.\\

%\begin{itemize}
%\item 
 \drule{\DNum(X),\DNum(U)}{\IfThen(X,X,U,U)}
%\item 
  \longdrule{\DNum(X_1),\DNum(X_2),\DNum(U_1),\DNum(U_2),\Neq(X_1,X_2)}{\IfThen(X_1,X_2,U_1,U_2)} 
%\end{itemize}

Thus, condition (1) can be guaranteed with respect to tgd $\sigma_1$
for all tuples $t_i$ by adding all atoms of the form $\IfThen(s_i,1,r_i,4)$ to \rchaseok.
 
Condition (2) is slightly more complicated. For our example tgd
$\sigma_1$ it says that if a tuple $t_i$ is obtained using $\sigma_1$
then the first tuple used for the chase step should be an
$R_2$-tuple. In terms of variables this can be stated as: if $s_i=1$
and $c_{i1}=j$ then $r_j=2$ (and likewise for the second atom of
$\sigma_1$. To express this IF-statement we use a 6-ary auxiliary
predicate $\IfTheni(X_1,X_2,Y_1,Y_2,U_1,U_2)$ expressing that if
$X_1=X_2$ and $Y_1=Y_2$ then $U_1=U_2$. It can be specified in \Pchase
by  the following three rules.\\

%\begin{itemize}
%\item 
  \drule{\DNum(X),\DNum(Y),\DNum(U)}{\IfTheni(X,X,Y,Y,U,U)}
%\end{itemize}

%\vspace{-10mm}

 \longdrule{\DNum(X_1),\DNum(X_2),\DNum(Y_1),\DNum(Y_2),\DNum(U_1),\DNum(U_2),\\\Neq(X_1,X_2)}{\IfTheni(X_1,X_2,Y_1,Y_2,U_1,U_2)}
 
\vspace{-15mm}

\longdrule{\DNum(X_1),\DNum(X_2),\DNum(Y_1),\DNum(Y_2),\DNum(U_1),\DNum(U_2),\\\Neq(Y_1,Y_2)}{\IfTheni(X_1,X_2,Y_1,Y_2,U_1,U_2)} 

For every pair of numbers $i,j\le N$, \rgoal then has atoms
$\IfTheni(s_i,1,c_{i1},j,r_j,2)$ and $\IfTheni(s_i,1,c_{i2},j,r_j,2)$.

In a similar fashion
\begin{itemize}
\item condition (3) yields one atom
$\IfThen(s_i,1,x_{i2},i)$, for every $i$;
\item condition (4) yields one atom
  $\IfThenii(s_i,1,c_{i1},j_1,c_{i2},j_2,x_{j_12},x_{j_21})$, for
  every $i,j_1,j_2\le N$, where \IfThenii is the 8-ary predicate for
  IfThen-statements with three conjuncts that can be defined
  analogously as \IfTheni;
\item condition (5) yields one atom
  $\IfTheni(s_i,1,c_{i1},j,x_{j1},x_{i1})$ for every $i,j\le N$.
\end{itemize}

Altogether, \rchaseok has $\bigO(N^3\ell k)$ atoms that together
guarantee that the variables of \rtuples encode an actual chase sequence.
}

\paragraph{\Pquery and \rquery.}
Finally, we explain how it can be checked that there is a homomorphism
from $q$ to $S$. We explain the issue through the little example query
$R_3(x,y)\land R_4(y,z)$. To evaluate this query, \rquery makes use of
two additional variables $q_1$ and $q_2$, one for each atom of
$q$. The intention is that these variables bind to the numbers of the
tuples that the atoms are mapped to. We have to make sure two kinds of
conditions. First, the tuples need to have the right relation symbol
and second, they have to obey value equalities induced by the
variables of $q$ that occur more than once.

\dlorfull{The first kind of conditions is checked by adding atoms
$\IfThen(q_1,i,r_i,3)$ and $\IfThen(q_2,i,r_i,4)$ to \rquery, for
every $i\le N$. The second condition is checked similarly. As we do not need any further auxiliary predicates, \Pquery is empty.
}
{
The first kind of conditions is checked by adding atoms
$\IfThen(q_1,i,r_i,3)$ and $\IfThen(q_2,i,r_i,4)$ to \rquery, for
every $i\le N$. The second kind of conditions can be checked by atoms 
$\IfTheni(q_1,i,q_2,j,x_{i2},x_{j1})$, for every $i,j\le N$.

As we do not need any further auxiliary predicates, \Pquery is empty
(but we kept it for symmetry reasons).

\smallskip
}

This completes the description of $P$.
Note that  $P$ is 
nonrecursive, and has polynomial size in the size of $q$ 
and $\Sigma$. In order to finish the proof
of part (a) of Theorem~\ref{theo:polyDatalog}, we next explain how to
reduce the arity of $P$.

This final step of the construction is based on two ideas.

First, by using Boolean variables and some new ternary relations, we
can replace the 6-ary relation $\IfTheni$ (and likewise the 4-ary
relation $\IfThen$). More precisely, we replace
every atom $\IfTheni(X_1,X_2,Y_1,Y_2,U_1,U_2)$ by a conjunction of the
form
\begin{multline*}
  \IfEq(X_1,X_2,B_1),\IfEq(Y_1,Y_2,B_2),\IfEq(U_1,U_2,B_3),\NotB(B_1,B_1'),\\
\NotB(B_2,B_2'),\OrB(B_1,',B_2',B_4),\OrB(B_3,B_4,B_5),\TrueB(B_5).
\end{multline*}

Here, $\NotB,\OrB,$ are predicates that mimic  Boolean gates,
e.g., $\OrB(B_3,B_4,B_5)$ holds if $B_5$ is the Boolean Or of $B_3$
and $B_4$, in particular all values have to be from
$\{0,1\}$. $\TrueB(B_5)$ only holds if $B_5=1$. The
predicate $\IfEq(X_1,X_2,B_1)$ holds if $B_1=1$ and $X_1=X_2$ or if
$B_1=0$ and $X_1\not=X_2$. The relations $\IfEq,\NotB,\OrB,\TrueB$ can
easily be defined in $\Pchase$.

The second idea is that $T$ need not be materialized. We only
materialize a
relation $T'$ of arity $a+1$ which is intended to represent all
database tuples. More precisely, $T'(j,X_1,\ldots,X_a)$ shall hold if
$(X_1,\ldots,X_a)$ represents a tuple from relation $R_j$ or if
$j=0$. Clearly, $T'$ can be defined in $\Ptuples$. 

Every tuple
$T(j,r_j,f_j,x_{j1},\ldots,x_{ja},s_j,c_{j1},\ldots,c_{jk})$ in
$\rtuples$ is then replaced by a conjunction of atoms with the same semantics.
The conjunct $T'(r'_j,x_{j1},\ldots,x_{ja})$ tests whether
$(x_{j1},\ldots,x_{ja})$ is in $R_{r'_j}$. Further atoms ensure that
$r_j=r'_j$ if $f_j=0$. Finally, it is ensured that, if $f_j=1$ the
values are restricted as by the right-hand side of rule \ref{eq:1}. \\

In order to prove part (b), we must get rid of the 
numeric domain (except for 0 and 1). This is actually very easy.
We just replace each numeric value by a logarithmic 
number of bits (coded by our 0 and 1 domain elements),
and extend the predicate arities accordingly. As a matter
of fact, this requires an increase of arity by a factor of 
$\log N = \bigO(\log |q|)$.  
\dlorfull{}
{
It is well-known that a successor predicate 
and a vectorized $<$ predicate for such bit-vectors
can be expressed by a 
polynomially-sized nonrecursive Datalog program, 
see~\cite{DEGV01}. The rest is completely 
analogous to the above proof.}
\dlorfull{
This concludes our explanation of the proof ideas underlying Theorem~\ref{theo:polyDatalog}. 
}
{
This concludes the proof sketch for Theorem~\ref{theo:polyDatalog}. 

\medskip

We would like to conclude this section with some remarks: 
}

\dlorfull{}{\smallskip}

\paragraph{Remark 1.} Note that the evaluation 
complexity of the Datalog program obtained for case (b) is
not significantly higher than the evaluation 
complexity of the program $P$ constructed for case (a). 
For example, in the most relevant case of bounded arities, both programs can be evaluated in NPTIME combined 
complexity over a database $D$. In fact, it is well-known that the combined complexity of a Datalog program of bounded arity is 
in NPTIME (see~\cite{DEGV01}). But it is easy to see that if 
we expand the signature of such a program (and of 
the underlying database) by a logarithmic number 
of Boolean-valued argument positions (attributes), nothing changes,
because the 
%number  joint 
possible values for such 
vectorized arguments are still of polynomial size. It is just 
a matter of coding. In a similar way, the data complexity in 
both cases (a) and (b) is the same (PTIME). 
  
\dlorfull{}{\smallskip}  
  
\paragraph{Remark 2.} It is easy to generalize this result to 
the setting where $q$ is actually a union of conjunctive queries (UCQ).

\dlorfull{}{

\smallskip
\paragraph{Remark 3.} The method easily generalizes to 
translate non-Boolean queries, i.e., queries with output,
to polynomially-sized nonrecursive Datalog programs with output. 
We are here only interested in {\em certain answers} consisting of tuples of values from the original domain $dom(D)$ (see \cite{FKMP05}).
Assume that the head of $q$ is an atom 
$R(X_1,\ldots,X_k)$ where $R$ is the output relation symbol, and the $X_i$  are variables also occurring in the body of $q$. 
We then 
obtain a nonrecursive Datalog translation by acting as in the 
above proof, except for the following modifications. 
Make  $R(X_1,\ldots,X_k)$  the head of 
rule \rgoal, and add for $1\leq i\leq$ an atom $adom(X_i)$
to \rquery, where $adom$ is an auxiliary predicate 
such that $adom(u)$  is iff  $u$ is in the {\em active non-numeric domain} of the database, that is, iff $u\in dom(D)$
and $u$ effectively occurs in the database. It is 
easy to see that the auxiliary predicate $adom$ itself can be 
achieved via a nonrecursive Datalog program from $D$. 
Clearly, by construction of (the so modified) program $P$, 
the output of $P$ are then precisely the certain answers of the query $q$.}
    
\dlorfull{}{

\smallskip
\paragraph{Remark 4.} The polynomially-sized nonrecursive 
Datalog program $P$ constructed in the proof of Theorem~\ref{theo:polyDatalog} can in turn be transformed in polynomial time into an equivalent first-order formula. 
In case of $N$-numerical databases this follows immediately from the constant depth of (the predicate dependency graph of) $P$.
Moreover, in case of non-numerical domains with two distinguished constants, the simulation of a numerical domain via bit-vectors can be easily expressed by a polynomially sized first-order formula. In summary, Theorem~\ref{theo:polyDatalog} remains valid if we replace {\em ``nonrecursive Datalog program''} 
by {\em ``first-order formula''}. However, for practical 
purposes nonrecursive Datalog may be the better choice, because 
the auxiliary relations that need to be computed only once 
are already factured out explicitly.}

\section{Further Results Derived From the Main Theorem}\label{sec:further}
We wish to mention some 
interesting consequences of Theorem~1 that 
follow easily from the above result after
combining it  with various other known results.

\subsection{Linear TGDs}

A linear tgd~\cite{CaGL09} is one that has a single atom 
in its rule body. The class of linear tgds is a 
fundamental one in the Datalog$^\pm$ family. 
This class contains the class of 
{\em inclusion dependencies}.
It was already shown in~\cite{JoKl84}
for inclusion dependencies that  
classes of linear tgds of bounded (predicate) arities 
enjoy the PWP.  That proof carries 
over to linear tgds\dlorfull{. }{, and we thus can state: 
 
\begin{lemma}
Classes of linear tgds of bounded arity 
enjoy the PWP. 
\end{lemma}}

By Theorem~\ref{theo:polyDatalog}, we then conclude:

\begin{theorem}\label{theo:linear}
Conjunctive queries under linear tgds of bounded arity 
are polynomially rewritable as nonrecursive Datalog 
programs in the same fashion as for Theorem~1. So are
sets of inclusion dependencies of bounded arity. 
\end{theorem}
 
\subsection{\bf DL-Lite}
A  pioneering and highly significant contribution 
towards tractable ontological reasoning was the introduction of
the \emph{DL-Lite} family of description logics (DLs) by Calvanese
et al.~\cite{CDLL*07,PLCD*08}. \emph{DL-Lite} was further studied and developed in~\cite{Za0}.

A DL-lite theory (or TBox) $\Sigma=(\Sigma^-,\Sigma^+)$ 
consists of a set of negative constraints 
$\Sigma^-$ such as key and disjointness constraints, and 
of a set $\Sigma^+$ of positive constraints that resemble 
tgds. As shown in~\cite{CDLL*07}, the negative constraints $\Sigma^-$ can be compiled into a polymomially sized 
first-order formula (actually a union of conjunctive queries) 
of the same arity as $\Sigma^-$ such that for each database 
and BCQ $q$, $(D,\Sigma)\models q$ iff $D\not\models\Sigma^-$ 
and $(D,\Sigma^+)\models q$. In (the full version of)~\cite{CaGL09} it was shown that for the main DL-Lite variants
defined in~\cite{CDLL*07}, each $\Sigma^+$ can be  
immediately translated into an equivalent set of linear 
tgds of arity 2. By virtue of this, and the above we obtain the following theorem.

\begin{theorem}\label{theo:A}
Let $q$ be a CQ and 
let $\Sigma=(\Sigma^-,\Sigma^+)$ be a DL-Lite theory 
expressed in one of the following DL-Lite variants: \textit{DL-Lite}$_{\mathcal{F},\sqcap}$, \textit{DL-Lite}$_{\mathcal{R},\sqcap}$, \textit{DL-Lite}$^+_{\A,\sqcap}$, 
\textit{DLR-Lite}$_{\mathcal{F},\sqcap}$, \textit{DLR-Lite}$_{\mathcal{R},\sqcap}$, or \textit{DLR-Lite}$^+_{\A,\sqcap}$.
Then $\Sigma^+$ can be rewritten into a 
nonrecursive Datalog program $P$ such that 
for each database $D$, $(D,\Sigma^+)\models q$
iff $D\models P$.  Regarding the arities of $P$, the same 
bounds as in Theorem~1 hold.
\end{theorem}

\subsection{\bf Sticky and Sticky Join TGDs}\label{subs:sticky}

Sticky tgds~\cite{CaGP10a} and sticky-join tgds~\cite{CaGP10a} 
are  special classes of tgds that generalize linear tgds 
but allow for a limited form of join (including as 
special case the cartesian product).  They allow one 
to express natural ontological relationships 
not expressible in DLs such as OWL. \dlorfull{For space reasons, we}{We} do not define these classes here, and refer the reader to~\cite{CaGP11}. By results of~\cite{CaGP11}, which will also be discussed in more detail in a future extended version\dlorfull{~\cite{extended}}{} of the present paper, both classes enjoy the Polynomial Witness Property. By Theorem~\ref{theo:polyDatalog}, we thus obtain the following result:  

\begin{theorem}
Conjunctive queries under sticky tgds and sticky-join tgds over a fixed signature
$\cal R$ are rewritable into 
polynomially sized nonrecursive Datalog programs of arity 
bounded as in Theorem~\ref{theo:polyDatalog}.
\end{theorem}

\section{Related Work on Query Rewriting}\label{sec:related}

Several techniques for query-rewriting have been developed.
An early algorithm, introduced in~\cite{CDLL*07} and implemented in
the QuOnto system\footnote{http://www.dis.uniroma1.it/~quonto/},
reformulates the given query into a union of CQs (UCQs) by means of
a backward-chaining resolution procedure. The size of the computed
rewriting increases exponentially w.r.t.~the number of atoms in the
given query. This is mainly due to the fact that unifications are
derived in a ``blind'' way from every unifiable pair of atoms, even
if the generated rule is superfluous.
An alternative resolution-based rewriting technique was proposed by
Per\'ez-Urbina et al.~\cite{JPMH09}, implemented in the Requiem
system\footnote{http://www.comlab.ox.ac.uk/projects/requiem/home.html},
that produces a UCQs as a rewriting which is, in general, smaller
(but still exponential in the number of atoms of the query) than the
one computed by QuOnto. This is achieved by avoiding many useless
unifications, and thus the generation of redundant rules due to such
unifications.  This algorithm works also for more
expressive non-first-order rewritable DLs. In this case, the
computed rewriting is a (recursive) Datalog query.
Following a more general approach, Cal\`i et al.~\cite{WCGP10}
proposed a backward-chaining rewriting algorithm for the first-order
rewritable Datalog$^\pm$ languages mentioned above. However, this
algorithm is inspired by the original QuOnto algorithm, and inherits
all its drawbacks. In ~\cite{GOP11}, a rewriting technique
for linear Datalog$^\pm$ into unions of conjunctive queries is proposed.
This algorithm is an improved
version of the one already presented in~\cite{WCGP10}\dlorfull{.}{, where further superfluous
unifications are avoided, and where, in addition, 
tedundant atoms in the body of a rule,
that are logically implied (w.r.t.~the ontological theory) by other
atoms in the same rule, are eliminated. This elimination of
body-atoms implies the avoidance of the construction of redundant
rules during the rewriting process.} However, the size of the
rewriting is still exponential in the number of query atoms. 

Of more interest to the present work 
are rewritings into 
nonrecursive Datalog.
In~\cite{Za1,Za2} a polynomial-size rewriting into nonrecursive Datalog is 
given for the description logics DL-Lite$^{\cal F}_{horn}$ 
and DL-Lite$_{horn}$. For DL-Lite$^{\cal N}_{horn}$, a DL with counting, 
a polynomial rewriting involving aggregate functions is proposed.
It is, moreover, shown in (the full version of)~\cite{Za1} that for the description logic 
DL-Lite$_{\cal F}$ a polynomial-size pure first-order 
query rewriting is possible. Note that neither of these
logics allows for role inclusion, while our approach covers description logics with role inclusion axioms.    
Other results in~\cite{Za1,Za2} are about {\em combined rewritings} where both the query and the database 
$D$ have to be rewritten. A recent very interesting paper discussing polynomial size 
rewritings is~\cite{Za3}. Among other results, \cite{Za3} provides complexity-theoretic arguments 
indicating that without the use of special constants (e.g, 0 and 1, or the numerical domain), a 
polynomial rewriting such as ours may not be possible.
Rosati et al.~\cite{AR10} recently proposed a very sophisticated
rewriting technique into nonrecursive Datalog, implemented in the
Presto system. 
This
algorithm produces a non-recursive Datalog program as a rewriting,
instead of a UCQs. This allows the ``hiding'' of the exponential
blow-up inside the rules instead of generating explicitly the
disjunctive normal form. The size of the final rewriting is, however,
exponential in the number 
of non-eliminable existential join
variables of the given query; such variables are a subset of the
join variables of the query, and are typically less than the number
of atoms in the query. Thus, the size of the rewriting is 
exponential in the query size in the worst case.
Relevant further optimizations of this method are given
in~\cite{OP11}.

\smallskip

\noindent{\bf Acknowledgment} 
G.~Gottlob's work was funded by the EPSRC Grant EP/H051511/1
ExODA: Integrating Description Logics and Database Technologies for Expressive Ontology-Based Data Access. We thank the anonymous referees, as well as Roman Kontchakov, Carsten Lutz, and Michael Zakharyaschev for useful comments on an earlier version of this paper.
\smallskip
\bibliographystyle{amsplain}
%\bibliography{DL11ABBREV}
\bibliography{DL11}
\end{document}